\pdfoutput=1

\documentclass[11pt]{article}

\usepackage{EMNLP2023}

\usepackage{multimedia}

\usepackage{subcaption}

\usepackage{microtype}
\usepackage{graphicx}
\usepackage{times}
\usepackage{latexsym}
\usepackage{listings}
\usepackage{xcolor}
\usepackage{lipsum}
\definecolor{codegreen}{rgb}{0,0.6,0}
\definecolor{codegray}{rgb}{0.5,0.5,0.5}
\definecolor{codepurple}{rgb}{0.58,0,0.82}
\definecolor{backcolour}{rgb}{0.95,0.95,0.92}

\lstdefinestyle{mystyle}{
    backgroundcolor=\color{backcolour},   
    commentstyle=\color{codegreen},
    keywordstyle=\color{magenta},
    numberstyle=\tiny\color{codegray},
    numbers=none,
    stringstyle=\color{codepurple},
    basicstyle=\ttfamily\footnotesize,
    breakatwhitespace=false,         
    breaklines=false,                 
    captionpos=b,                    
    keepspaces=true,                 
    numbers=left,                    
    numbersep=5pt,                  
    showspaces=false,                
    showstringspaces=false,
    showtabs=false,                  
    tabsize=2
}
\usepackage[T1]{fontenc}

\usepackage[utf8]{inputenc}

\usepackage{microtype}
\usepackage{booktabs}
\usepackage{graphicx}
\usepackage{amsmath}

\usepackage{inconsolata}

\newcommand\icon{\raisebox{-3.0pt}{\includegraphics[width=2.0em]{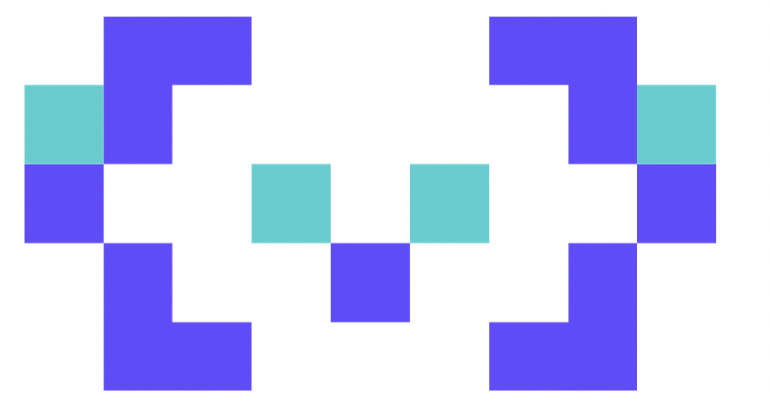}}}
\newcommand{\modelname}{ModelScope-Agent }

\title{\icon ModelScope-Agent: Building Your Customizable Agent System with Open-source Large Language Models}

\author{
Chenliang Li, Hehong Chen, Ming Yan\thanks{\ \ Corresponding author: <ym119608@alibaba-inc.com>}, Weizhou Shen, Haiyang Xu, Zhikai Wu \\
  \textbf{Zhicheng Zhang, Wenmeng Zhou, Yingda Chen, Chen Cheng, Hongzhu Shi} \\
  \textbf{Ji Zhang, Fei Huang, Jingren Zhou} \\
  DAMO Academy, Alibaba Group, China \\
  \\
}

\begin{document}
\maketitle
\begin{abstract}

%
Large language models (LLMs) have recently demonstrated remarkable capabilities to comprehend human intentions, engage in reasoning, and design planning-like behavior. To further unleash the power of LLMs to accomplish complex tasks, there is a growing trend to build agent framework that equips LLMs, such as ChatGPT, with tool-use abilities to connect with massive external APIs.

In this work, we introduce ModelScope-Agent, a general and customizable agent framework for real-world applications, based on open-source LLMs as controllers. It provides a user-friendly system library, with customizable engine design to support model training on multiple open-source LLMs, while also enabling seamless integration with both model APIs and common APIs in a unified way. To equip the LLMs with tool-use abilities, a comprehensive framework has been proposed spanning over tool-use data collection, tool retrieval, tool registration, memory control, customized model training, and evaluation for practical real-world applications. Finally, we showcase ModelScopeGPT, a real-world intelligent assistant of ModelScope Community based on the \modelname framework, which is able to connect open-source LLMs with more than 1000 public AI models and localized community knowledge in ModelScope. The \modelname library\footnote{https://github.com/modelscope/modelscope-agent} and online demo\footnote{https://modelscope.cn/studios/damo/ModelScopeGPT/summary} are now publicly available.


\end{abstract}

\section{Introduction}
Large language models~\cite{chatgpt,gpt4,touvron2023llama,palm} have gradually become common AI assistants that demonstrate great potential in comprehending human intentions, performing complex reasoning tasks, and enabling content creation. Despite the rapid advancements of open-source LLMs, e.g., LLaMA~\cite{touvron2023llama} and ChatGLM~\cite{chatglm}, they still remain limited in performing complex tasks, such as following user instructions to use external tools and capture up-to-date information. 

To further unleash the power of LLMs for real-world practical applications, a rising trend of current research~\cite{schick2023toolformer,hugginggpt,gpt4tools,qin2023tool,gorilla} begins to enable LLMs with tool-use abilities towards building an AI Agent. These include HuggingGPT~\citep{hugginggpt}, Visual-ChatGPT~\citep{visualchatgpt} and Gorilla~\citep{gorilla} for connecting with HuggingFace models, ToolAlpaca~\citep{toolalpaca} and ToolLLaMA~\citep{qin2023tool} for using massive common APIs such as weather forecast and search engine. These methods either directly rely on closed-source counterparts like ChatGPT or focus on certain types of API tools. Recently, there have also been public releases of AI agents, such as Auto-GPT\footnote{\small{https://github.com/Significant-Gravitas/Auto-GPT}}, LangChain\footnote{\small{https://github.com/langchain-ai/langchain}} and Transformers Agent~\citep{Transformers_agent}, which enable LLMs, such as ChatGPT or GPT-4, to use tools and solve complex AI tasks. However, these agents are mainly built with closed-source LLMs and how to build a customizable agent system with open-source LLMs remains largely unexplored.



In this work, we present ModelScope-Agent, a general and customizable agent system for real-world applications, based on open-source LLMs as controllers. ModelScope\footnote{\small{https://modelscope.cn/models}} is a public ML community, which seeks to bring together the most advanced machine learning models from the AI community, and streamlines the process of leveraging AI models in real-world applications. \modelname provides a flexible and user-friendly system library, with customizable engine design to support model training on multiple open-source LLMs, while also enabling seamless integration with both model APIs and common APIs in a unified way. It features an LLM-centric system design, which includes open-source LLMs as core controller, and further interact with a tool-use module and a memory module to accomplish complex tasks. At the core of \modelname, the library supports flexible selection and training on various open-source LLMs, such as LLaMA~\cite{touvron2023llama}, ChatGLM~\cite{chatglm}, ChatPLUG~\cite{tian2023chatplug} and other customized LLMs in ModelScope. For tool use, \modelname provides a default tool library, which supports diverse AI model APIs across NLP, CV, Audio and Multi-model fields, as well as massive common APIs such as search engine. It also supports registering new self-defined API plugins and automatic API retrieval from the large tool library. It is easy for users to customize their most appropriate LLMs, local API tools and functions to develop real-world applications. Moreover, a memory module is also introduced to better store and manage the system message, user history, in-context examples, tool message and localized knowledge.

To enable the open-source LLMs to better control the whole agent system, we further propose a comprehensive framework of tool-use data collection, customized model training, evaluation and deployment. Notably, we release a comprehensive  tool-enhanced dataset MSAgent-Bench, which consists of 598k dialogues with various API categories, multi-turn API calls, API-Oriented QA, and API-Agnostic instructions in both English and Chinese. A simple training strategy of Weighted LM, that enhances the training of generation of API name and parameters, is used to better ensure the correctness of API calls. Besides, an evaluation framework is also supported in our library to examine the tool-use abilities of the trained models in different aspects. Furthermore, we applied \modelname in a real-world application of ModelScope Community namely ModelScopeGPT, which is able to connect open-source LLMs with more than 1000 public AI models and access localized community knowledge in ModelScope for community QA.


To summarize, \modelname is a general and customizable agent system designed for developers to harness the power of open-source LLMs. The library targets the following goals:

\begin{itemize}

\item[\textbullet] Agent based on Open-Source LLMs: the controller of \modelname can be flexibly selected from open-source LLMs that are optimized through our agent training framework.

\item[\textbullet] Support and Customization of Diverse Tools: Dozens of diverse model APIs and common APIs are given by default. The library supports registering new self-defined APIs and automatic API retrieval from the toolset.

\item[\textbullet] Customizable of Applications: \modelname can be flexibly applied in various industry applications. The agent and training framework are documented describing its usage, construction and optimization.

\end{itemize}

\modelname is in continual development by the engineers at ModelScope and is released under an Apache 2.0 license.  Full documentation is available through the project website.



\begin{figure*}[h]
     \centering
     \includegraphics[width=\linewidth]{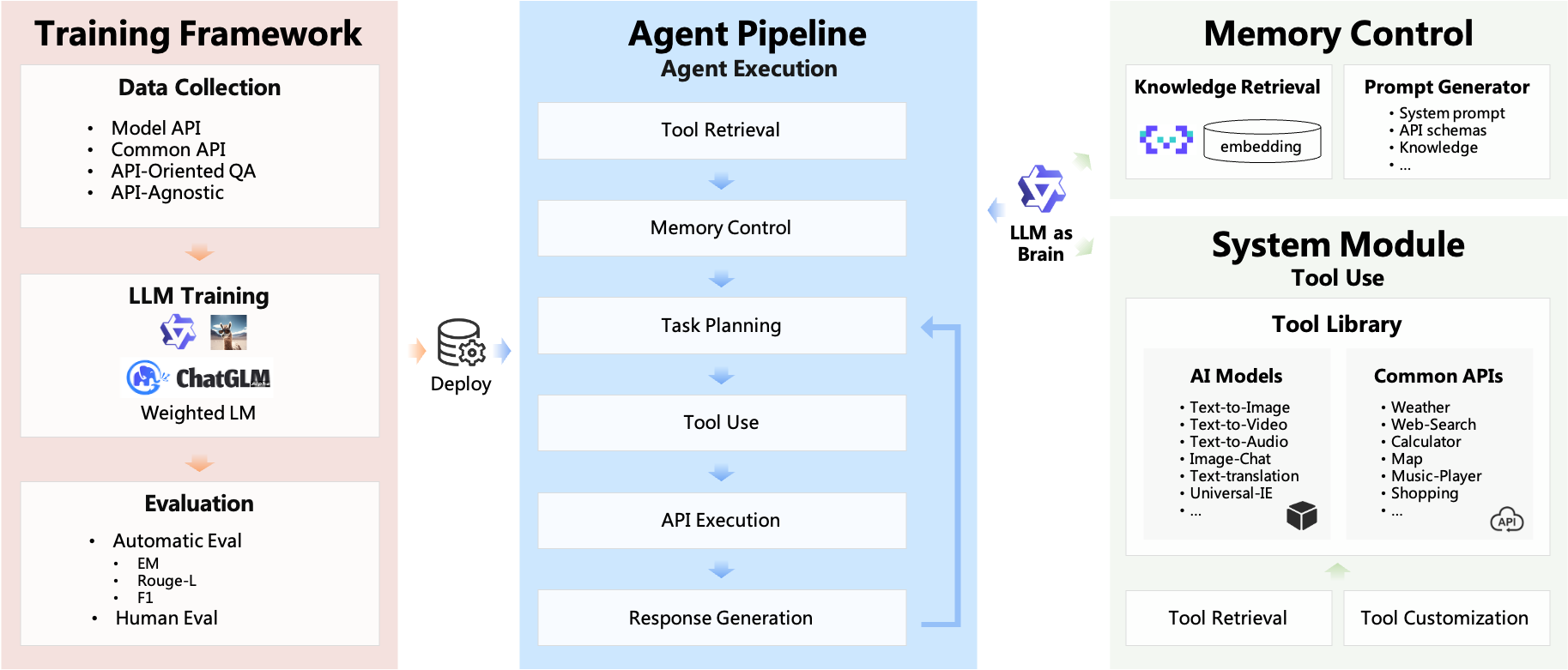}
     \caption{The overall system architecture of ModelScope-Agent.}
     \label{fig:framework}
     \vspace{-0.5cm}
\end{figure*}

\section{The ModelScope Agent} 


\modelname is designed to facilitate developers in building customizable agent systems based on open-source LLMs. The overall system architecture is shown in Figure~\ref{fig:framework}. It includes open-source LLMs as controller, a tool-use module and a memory module to interact with. Given human instruction, the Agent, which adopts the selected LLM as the controller, will automatically plan tasks, selectively uses tools, leverage knowledge in memory, and finally provides helpful responses to users.  

\subsection{LLMs as Brain}
LLMs serve as the brain of the agent, responsible for planning and decomposing user requests, selectively calling tools, performing retrieval, and integrating all the information from previous steps to generate the final response. In order to make it easier for users to customize the agent with their own LLMs, we have added support for various open-source LLMs by default, such as LLaMA, ChatGLM and ChatPLUG, which have been optimized through our tool learning pipeline. The details of training strategy and tool-use datasets can be referred to Section \ref{training}. \modelname has integrated the LLM inference pipeline of the ModelScope community, and replacing LLMs can be done by simply setting the \textit{model\_name} and \textit{model\_config}. In \textit{model\_config}, the model\_id, model\_revision, and model parameter settings such as max sequence length, should be configured.
\begin{lstlisting}[language=python,numbers=none]
# LLM config "cfg_file"
from modelscope.utils.config import Config
model_cfg = Config.from_file(cfg_file)
llm = LocalLLM(model_name, model_cfg)
\end{lstlisting}

Furthermore, the \modelname also provides a standard way to integrate new LLM. Users can add their own LLMs, by integrating the LLM pipeline into ModelScope. After that, the agent can select the new LLMs for training and inference. 





\subsection{Tool Use}

\paragraph{Tool Library}
%
The tool library is used to configure and manage various collections of APIs used in the agent. \modelname can support a wide range of both common APIs such as search APIs, and AI model APIs across NLP, CV, Audio and Multi-modal models in ModelScope and HuggingFace. Each tool API consists of the API name, description, parameters and request functions. Users can easily choose and configure proper APIs in the library to build their own agent. The default APIs supported in the library can be referred to Appendix~\ref{sec:tool_list}. 
\begin{lstlisting}[language=python,numbers=none]
# tool default config file "default_file"
tool_cfg = Config.from_file(default_file)
\end{lstlisting}

\paragraph{Register and Customize New Tool} 
The agent allows users to register and customize new tools, while also supporting quick integration of newly registered tools into the agent, enabling LLMs to selectively use the additional self-defined tools for specific applications. This can be simply done by inheriting from a base class, namely \textit{Tool}, and defining a new \textit{CustomTool} with the API-related schema of API name, description, parameters, and request functions. More details about \textit{CustomTool} can be referred in Appendix \ref{sec_custom}.  

\begin{lstlisting}[language=python,numbers=none]
from modelscope_agent.tools import Tool
class CustomTool(Tool):
     # logic added here
     # refer example in Appendix A.2
tool_list = {'customo-tool': CustomTool()}
\end{lstlisting}

\paragraph{Tool Retrieval and Execution}

Due to the large amount of tool APIs in the tool library, a tool retrieval module is further introduced to recommend appropriate APIs for each instruction prompt. Specifically, we use the dense vector retrieval method based on the unified multilingual text-embedding API~\footnote{https://help.aliyun.com/zh/dashscope/getting-started-1}. We vectorize both the text descriptions of the APIs and the instruction prompt using the text-embedding API. The top-3 most relevant APIs with the highest vector product scores are selected for tool use. As a result, the schema information of the retrieved APIs will be concatenated with other system prompts in the subsequent memory module and sent to LLMs as input. With the concatenated instruction prompt, the LLMs will plan and generate the API request, which will be executed by the agent. The agent will then return the results to the LLMs for continuous generation.


\subsection{Memory Control}
The memory module is used to retrieve, and assemble a series of contextual information as input to the LLMs. It consists of a knowledge retrieval submodule and a prompt generator submodule, which are responsible for external knowledge retrieval and instruction prompt generation, respectively.

\paragraph{Knowledge Retrieval} It enables the agent to get access to up-to-date and localized information related with query prompt, thereby augmenting LLMs with dynamic and domain-specific knowledge. We follow the same dense vector retrieval method as the previous tool retrieval module, and support large-scale knowledge retrieval from localized document corpus. Similarly, it allows users to customize by changing to other open-source retrieval frameworks. 



\paragraph{Prompt Generator} The prompt generator is used to assemble all available contextual information such as system prompt, API schema, retrieved knowledge, conversation history, and few-shot examples. According to the type of user query and the maximum length of the LLM, the users can selectively choose proper contextual information and assemble the required input to the LLM. In our agent, the prompt generator needs to be defined before the agent is constructed.


\subsection{Agent Pipeline}

In summary, we build the agent by combining all the modules: LLM controller, tool-use module, and memory module. With \textit{agent.run}, the agent can efficiently execute and complete the instruction in a one-step generation. First, the agent retrieves query-related tools through the tool retrieval and combines the retrieved API schema with other contextual prompts in memory module, to construct a new instruction prompt. Then, the agent sends this new prompt to the LLM, who plans whether and which API to call and generate an API request. Next, the agent will execute the selected API with the extracted API parameters and return the API results to the LLMs, which will continue to plan whether to call other APIs. If another API call is needed, the process is repeated, otherwise, the LLMs generate the final response and the agent returns the final result to the user.
\begin{lstlisting}[language=python,numbers=none]
agent = AgentExecutor(llm, tool_cfg,        
          additional_tool_list=tool_list)
agent.run("Draw a logo image of agent")
\end{lstlisting}

\begin{table*}[t]
 \centering
	\resizebox{0.999\textwidth}{!}{
	\begin{tabular}{l|cccccc}
		\toprule
		\textbf{Dataset}  & \textbf{Language} &\textbf{Instance Type} & \textbf{\# Instances} &\textbf{API type} & \textbf{Avg. Turn} &\textbf{Avg. Step} \\ 
            \hline
            API-Bank~\citep{apibench} & English & Tool Use& 264 & Common API & 3.27 & 1.92\\
            ToolAlpaca~\citep{toolalpaca} & English & Tool Use & 3.9 K& Common API & 1 & 1.66 \\
            Gorilla~\citep{gorilla} & English & Tool Use & 16.4 k & Model API& 1 & 1\\
            GPT4Tools~\citep{gpt4tools} & English & Tool Use & 71.4 K & Model API & 1 & 1\\
            ToolBench~\citep{qin2023tool} & English & Tool Use & 26.9 K & Common API & 1 & 4.1 \\
            \hline
            MSAgent-Bench (ours) & English + Chinese & Tool Use + Common Chat & 598 K & Common API + Model API & 1.52 & 1.31 \\
            \bottomrule
	\end{tabular}
	}
        \caption{The statistics of MSAgent-Bench and other existing tool learning datasets.}
	\label{tab:compare_dataset}
\end{table*}

\begin{figure*}[t]
    \centering
    \includegraphics[width=0.9\textwidth]{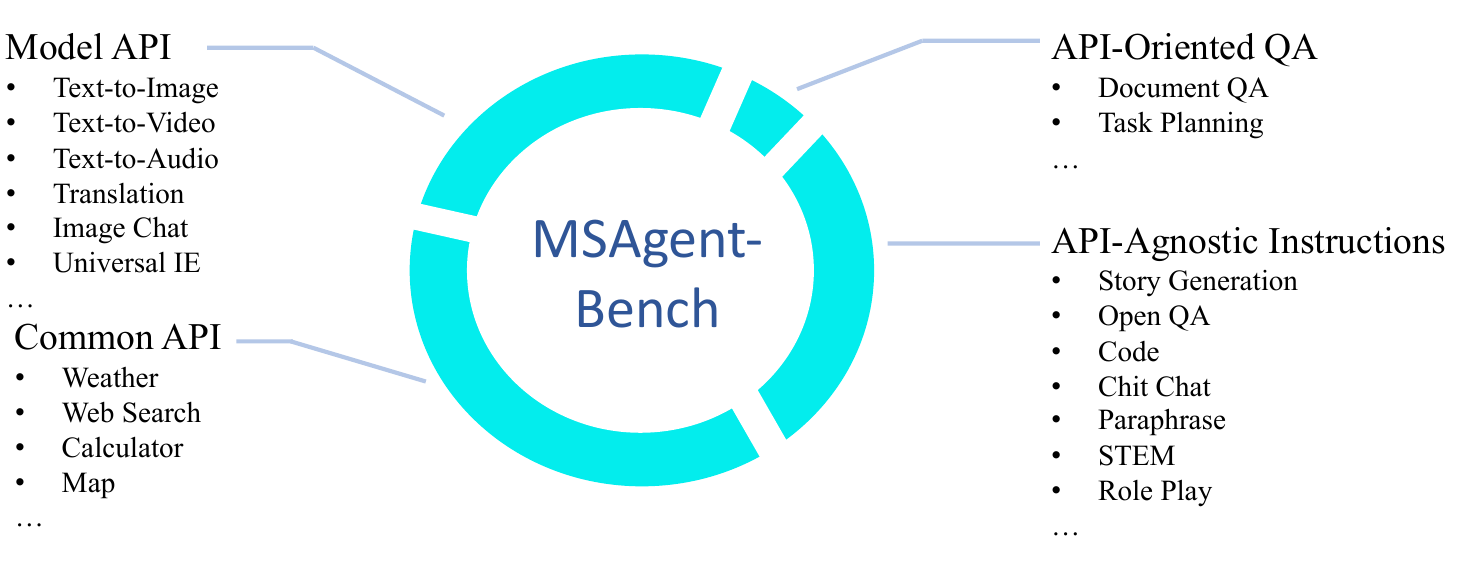}
	\caption{The instance types and distribution of our collected MSAgent-Bench.}
	\label{fig:data_collection}
	\vspace{-0.3cm}
\end{figure*}

\section{Training}
\label{training}


\subsection{Dataset}\label{sec:dataset}





To facilitate building an agent with the ability to use tools while upholding an optimal level of user engagement, we release a comprehensive tool dataset, MSAgent-Bench\footnote{https://modelscope.cn/datasets/damo/MSAgent-Bench/summary}, utilizing ChatGPT synthetic data and the existing instruction-following datasets. Our released dataset encompasses 598k dialogues. Table~\ref{tab:compare_dataset} outlines the key differences between the released dataset and other public available tool learning datasets, while the data distribution of our dataset is illustrated in Figure~\ref{fig:data_collection}. As demonstrated in the Table and Figure, we have made certain efforts to construct a comprehensive dataset which enables the effective training of an agent:

\noindent\textbf{Multilingual:} We collect instances in both Chinese and English, ensuring that the trained agent is capable of functioning in both languages.

\noindent\textbf{Various API Categories:} Our dataset supports Common APIs that have been registered by users or applied through online API platforms, as well as model APIs that can call neural models.

\noindent\textbf{Multi Turn Dialog:}  In real-life scenarios, agents may need to request more specific clarification from users to complete a task or receive additional instructions after completing a previous task. Our dataset accounts for these scenarios and supports multi-turn user-agent interactions when using tools.

\noindent\textbf{API-Oriented QA:}  An effective agent should possess knowledge of APIs. Our dataset incorporates API document QA tasks and task planning tasks which requires agents to offer appropriate suggestions to users on how to use various APIs to solve complex tasks.

\noindent\textbf{API-Agnostic Instructions:} To enhance the agent's ability to follow common instructions and increase user engagement, we have incorporated both Chinese and English API-agnostic instructions within our dataset. These instructions place greater emphasis on the agent's inherent capabilities rather than reliance on API invocation.

The data was collected by prompting ChatGPT (\texttt{gpt-3.5-turbo}) to generate instructions, API requests, and answers based on the API calling results, more details can be accessed in Appendix~\ref{app:data_collect}.

\subsection{Model Training}

\begin{figure*}[h]
     \centering

     \begin{subfigure}{0.35\textwidth}
         \centering
         \includegraphics[width=\textwidth]{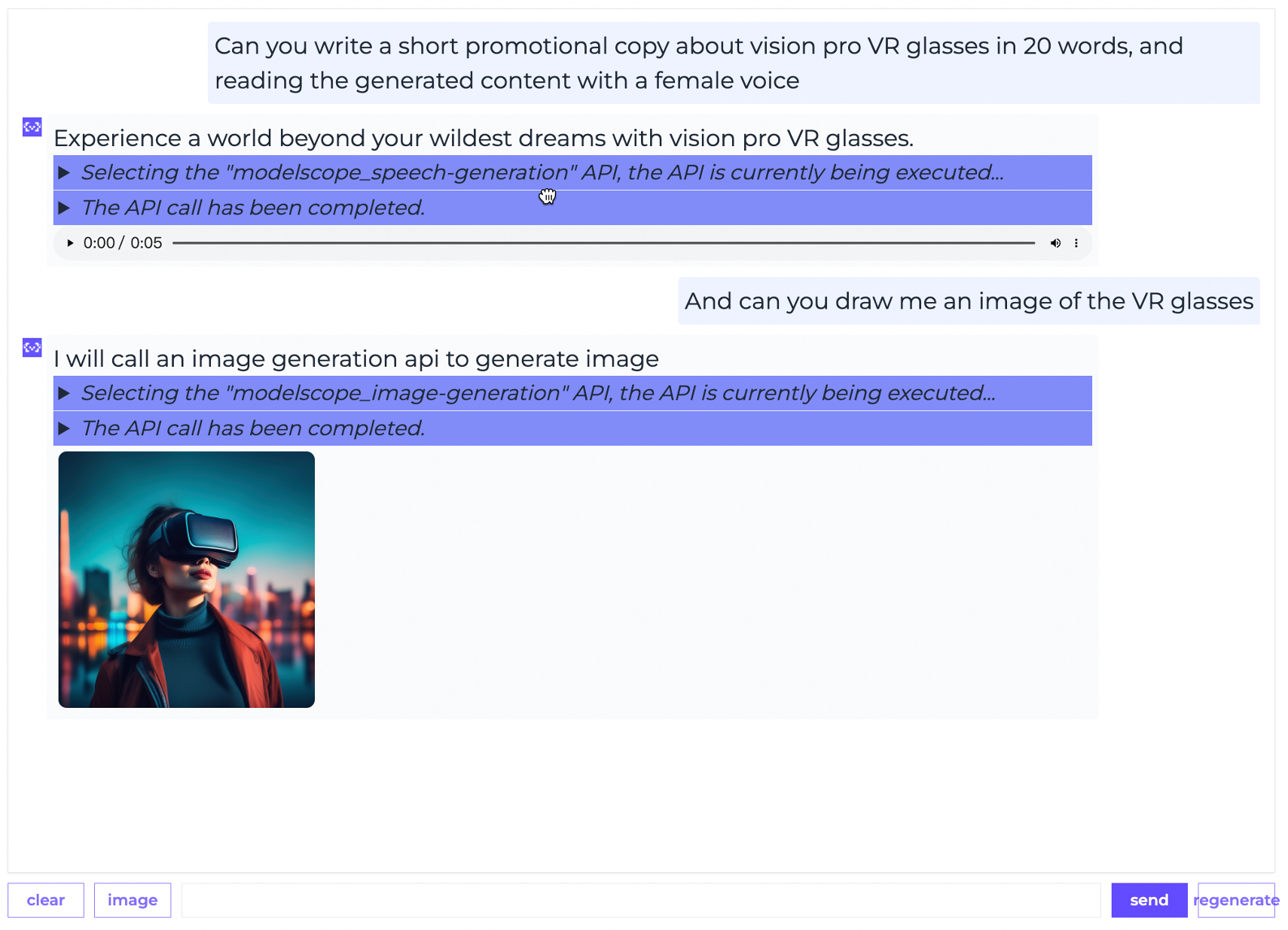}
         \caption{ModelScope Intelligent Assistant}
         \label{fig:form1}
     \end{subfigure}
     \hfill
     \begin{subfigure}{0.6\textwidth}
         \centering
         \includegraphics[width=\textwidth]{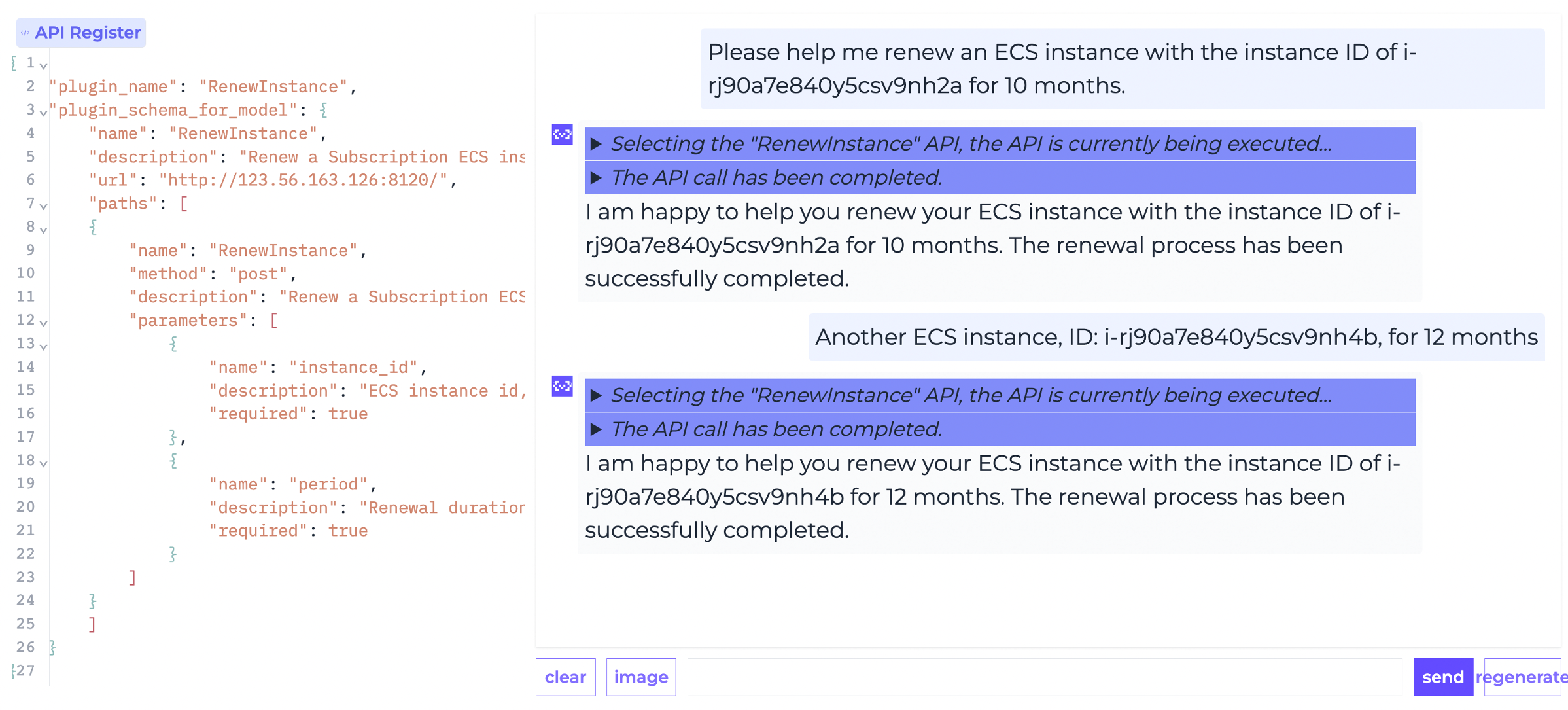}
         \caption{Register and Use New Tools on Alibaba Cloud}
         \label{fig:table1}
     \end{subfigure}
    \caption{Demo cases of ModelScopeGPT based on \modelname.}
\end{figure*}


We use the MSAgent-Bench to fine-tune multiple open-source LLMs, including LLaMA~\cite{touvron2023llama}, Qwen~\citep{qwen}, ChatPLUG~\cite{tian2023chatplug} etc. We train all the open-source LLMs in a multi-round conversation mode and concatenate all the prompts and answers. Compared to common instruction tuning data, the tool learning samples focus more heavily on the accuracy of tool selection and API parameter prediction. Therefore, we propose a simple training strategy, Weighted LM, which enhances the training of generation of API name and parameters, while zero-out the loss of tokens from the user prompt and the tool execution. More details can be be referred to Appendix \ref{sec_weighted_lm}.  

\begin{lstlisting}[language=python,numbers=none]
kwargs = dict(model=model, ...)
trainer: EpochBasedTrainer = build_trainer
   (name=args.trainer, default_args=kwargs)
trainer.train()
\end{lstlisting}







\section{Evaluation} 
\label{evaluation}
Our evaluation system, MSAgent-Eval, comprises two modules: an automatic evaluation framework which comprehensively evaluates API usability of the agents, and a human evaluation framework implemented by an agent arena which reflects the preferences of human users. 

\subsection{Automatic Evaluation Framework}

In automatic evaluation, we mainly focus on evaluating agent's ability to generate accurate API request and the proper answers according to the API calling results. Specifically, we use the action exactly match score (Action EM) which measures whether the agent uses the correct API as the reference gold API, and the ROUGE-L score which measures the similarity between the generated response and the gold answer. Additionally, we introduce a novel metric called Argument F1 for fully evaluating the quality of API requests. To compute Argument F1, we categorize the arguments in agent's API request  into two cases, namely Half match (HM) and Full match (FM), representing correct argument but with wrong value and correct argument with correct value, respectively. Suppose the gold argument number in the API is $|A|$, and the number of arguments in the agents API request is $|A^*|$, we compute the new Recall and Precision as follows:
\begin{align}
    &R = (0.5\times \text{\# HM} + \text{\# FM})/{|A|} \\
    &P = (0.5\times \text{\# HM} + \text{\# FM})/{|A^*|}
\end{align}
and the final argument F1 is computed as:
\begin{equation}
    F1 = 2(R*P)/(R+P).
\end{equation}

A sample code for the automated evaluation of agents is provided below:
\begin{lstlisting}[language=python,numbers=none]
from tool_agent_finetune import evaluation
EM, F1, ROUGE = evaluation(refs, preds)
\end{lstlisting}
Expert annotators were engaged to annotate the evaluation instances, with the task of providing diverse instructions, manually documenting correct API calling requests, and writing appropriate responses. The statistics  of our currently assembled test data is in Appendix~\ref{app:testset}, and the automatic evaluation scores of our trained agents can be found in Appendix~\ref{app:evaluate_result}. We also guarantee the users to upload their own annotated test examples to accurately evaluate the performance of agents in customized scenarios.


\subsection{Human Evaluation with Agent Arena}
Inspired by the Arena for ChatBots~\citep{arena}, we have built an accessible Agent Arena~\footnote{https://modelscope.cn/studios/LLMZOO/Chinese-Arena/summary} that allows users to furnish instructions to two anonymous agents, based on the provided APIs. Subsequently, users have the opportunity to vote on which Agent performs better in tackling the instruction with the given APIs. In accordance with the framework presented by~\citet{arena}, we adopt a system of ELO ratings and leaderboard maintenance for the participating Agents.

\section{Usage Example of ModelScopeGPT}
In this section, we showcase a successful application of ModelScope Community, ModelScopeGPT\footnote{\small{https://modelscope.cn/studios/damo/ModelScopeGPT\\/summary}}, based on our ModelScope-Agent.

\paragraph{ModelScope Intelligent Assistant} Based on  \modelname, we have developed an intelligent assistant for the ModelScope Community, namely ModelScopeGPT. It uses LLMs as a controller to connect dozens of domain-specific AI models in the ModelScope open-source community, covering NLP, CV, Audio, and Multi-Modal fields. To make the pipeline more practical, we have included API retrieval and knowledge retrieval tool to automatically select proper APIs and get access to the local ModelScope knowledge. As shown in Figure \ref{fig:form1}, ModelScopeGPT can support API calls in multi-turn conversations and generate correct API call parameters using information from previous conversations. More cases can refer to Appendix \ref{cases}. As a result, ModelScopeGPT has achieved a total request number of over 170k from 40k user visits within one month after its release. 

\paragraph{Register and Use New Tools} Another key feature of an agent is its generalization capability to unseen APIs. This allows users to quickly register their own APIs and customize their specific applications. Therefore, we test the generalization ability of ModelScopeGPT by applying it to an Alibaba Cloud application scenario. As shown in Figure~\ref{fig:table1}, we first found an API for renewing an ECS instance on Alibaba Cloud. Then, we registered the API schema defined in the tool library to the agent. Finally, we entered the prompt "\textit{Please help me renew an ECS...}" in the demo. The agent generated a request through planning, selected the appropriate API, called the API to renew the instance successfully, and provided a reply to inform the user that the renewal was completed. This test demonstrates that the open-source LLM optimized based on the released API dataset has a strong generalization ability towards unseen APIs.





\section{Conclusion}
\modelname aims to facilitate building AI Agent applications and research based on open-source LLMs by providing a general and customizable agent framework covering flexible system design, data collection, model training, evaluation and usage example in real-world application. It provides an open-source, community-driven library towards AI Agent learning and best practices for building an agent system with open-source LLMs. We hope \modelname can help pave the way towards a new era of AI Agent. 


\section*{Ethics Statement}
\paragraph{Intended Use.} \modelname is designed to facilitate building AI Agent applications and research based on open-source LLMs, by providing a general and customizable agent system. 

\paragraph{Potential Misuse.} Although we have only trained with the tool-use datasets and gone through certain data filtering rules, it is still possible that the customized model may generate some biased, fake, and unsafe information. Our agent framework also provides users with the freedom to select proper LLMs and upload their own clean data for training. It is also important to design specific methods to improve the safety of the agent framework in the future. 



\bibliography{arxiv}
\bibliographystyle{acl_natbib}

\clearpage

\appendix

\section{Library}
\label{library}

\subsection{Tool List}
\label{sec:tool_list}
\begin{table}[h]
 \centering
 \tiny
{
	\begin{tabular}{l|c|c}
		\toprule
		\textbf{API Name (language)} &  \textbf{Description}  & \textbf{Type} \\ \hline 
         Text-to-Image(en)   & Converts text to an image. & Model API   \\
         Text-to-Image(zh)   & Converts text to an image. & Model API   \\         
         Text-to-Video(en)  & Converts text to a video. & Model API\\
         Text-to-Audio(en) & Converts text to audio. & Model API\\
         Text-to-Audio(zh) & Converts text to audio. & Model API\\
         Image-Chat(en) & Image chat. & Model API \\
         Translation-zh2en & Translates Chinese text to English. & Model API  \\
         Translation-en2zh & Translates English text to Chinese. & Model API \\
         Universal-IE(zh) & Extracts structured information. & Model API \\
         Text-to-Geographic(zh) & Extracts geographic information. & Model API \\
         NER(zh) & Recognizes named entities in text. & Model API \\
         API-Retrieval & Retrieves relevant APIs & Common API \\
         ModelScope-Retrieval & Retrieves modelscope docs. & Common API \\
            \bottomrule
	\end{tabular}
	}
        \caption{The statistics of default tool list. Supported input languages for the APIs are listed in parentheses.}
	\label{tab:tool_list}
 \vspace{-0.3cm}
\end{table}

\subsection{CustomTool}
\label{sec_custom}
User can customize their own tools by inheriting a base tool and defining the tool names, descriptions, and parameters according to a pre-defined schema. Moreover, you can implement \textit{\_local\_call()} or \textit{\_remote\_call()} depending on your specific requirements. To illustrate, below is an example of a custom tool:

\begin{lstlisting}[language=python,numbers=none]


class CustomTool(Tool):
    description = 'xxx'
    name = 'xxx'
    parameters: list = [{
        'name': 'xxx',
        'description': 'xxx',
        'required': True
    }]

    def _local_call():
        ...

    def _remote_call():
        ...
\end{lstlisting}

\section{Experiment Setup}
\label{sec:training_config}

\subsection{Evaluation Benchmark}\label{app:testset}
To assess the generalization of the trained agent, we include 10 in-domain APIs that appear in the training set of ModelScope-Agent and 10 real unseen APIs\footnote{In progress, we will include more APIs in the future.}. We also account for the multi-turn ability of the agent by annotating several multi-turn scenarios in our evaluation benchmark. Our test instances were annotated by asking the human experts to write diverse instructions first. Then the human experts were ask to write the JSON API request and answer the instructions properly after obtaining the API calling results. Our final testing dataset consisted of 360 conversations with 2059 text snippets as the references to be compared with the agent prediction, which comprise 798 API requsts and 1261 plain text answers according to the previous calling results.

\subsection{Evaluation Results}\label{app:evaluate_result}
\begin{table}[h]
 \centering
	\resizebox{0.78\linewidth}{!}{
	\begin{tabular}{lccc}
		\toprule
		
		\textbf{Model} & \textbf{ROUGE-L}&  \textbf{Action EM}& \textbf{Argument F1}\\ \hline \hline
             ChatGPT (2-shot)$^*$ & 36.70& 34.82 & 25.51\\
             LLaMA  & 39.16&58.60 &44.98 \\
            ChatPLUG\footnote{https://modelscope.cn/models/damo/ChatPLUG-3.7B/summary}   & 46.45& 68.29&  55.12\\
             MSAgent-Qwen\footnote{https://modelscope.cn/models/damo/MSAgent-Qwen-7B/summary}  & 51.35 & 87.23 & 68.09 \\
            \bottomrule
	\end{tabular}
	}
        \caption{Automatic evaluation results. $^*$ represents that we do not fine-tune ChatGPT but use in-context learning with 2 demonstrations.}
	\label{tab:results}
 \vspace{-0.3cm}
\end{table}
We compare the models trained in our proposed ModelScopeGPT. The automaction evaluation results are shown in Table~\ref{tab:results}. Based on the findings obtained from our experimentation, it is evident that ChatGPT with in-context learning yielded inferior results as compared to other models that were subjected to finetuning. Furthermore, LLaMA underperformed when compared to other finetuned models. Our error study revealed that the lower performance of ChatGPT and LLaMA could be attributed to a large proportion of Chinese test cases in our test set. The models (ChatPLUG, Qwen) that performed better were those that predominantly focused on Chinese data. Our investigation revealed that ChatGPT and LLaMA exhibited limitations in user intent recognition, which ultimately led to their suboptimal performance on Action EM. Among the models examined, Qwen displayed the most favorable performance, which could be attributed to the superior performance of its basic model.

\begin{figure*}[h]
     \centering
     \includegraphics[width=\linewidth]{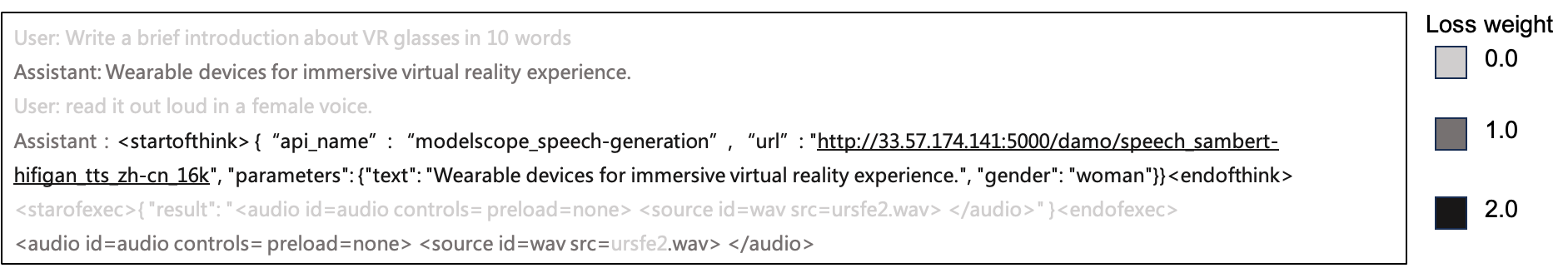}
     \caption{Example of training strategy for weighted LM. Different colored tokens have different loss weights.}
     \label{fig: weightedlm}
\end{figure*}

\begin{figure*}[h]
     \centering
     \includegraphics[width=\linewidth]{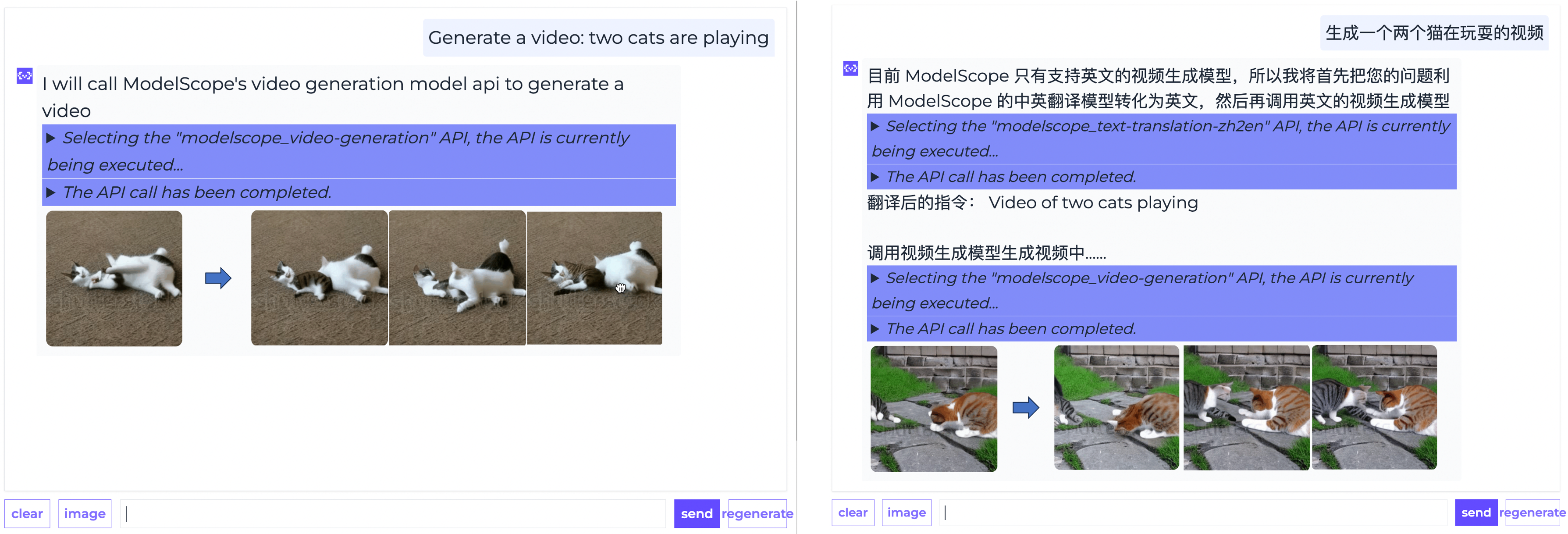}
     \caption{Single-step tool-use instructions, text-to-video cases. We have captured a few frames of the video to display. Testing the model using the same semantic instruction in both English (left) and Chinese (right).}
     \label{fig: case_video_generation}
\end{figure*}

\begin{figure*}[!htb]
     \centering
     \includegraphics[width=\linewidth]{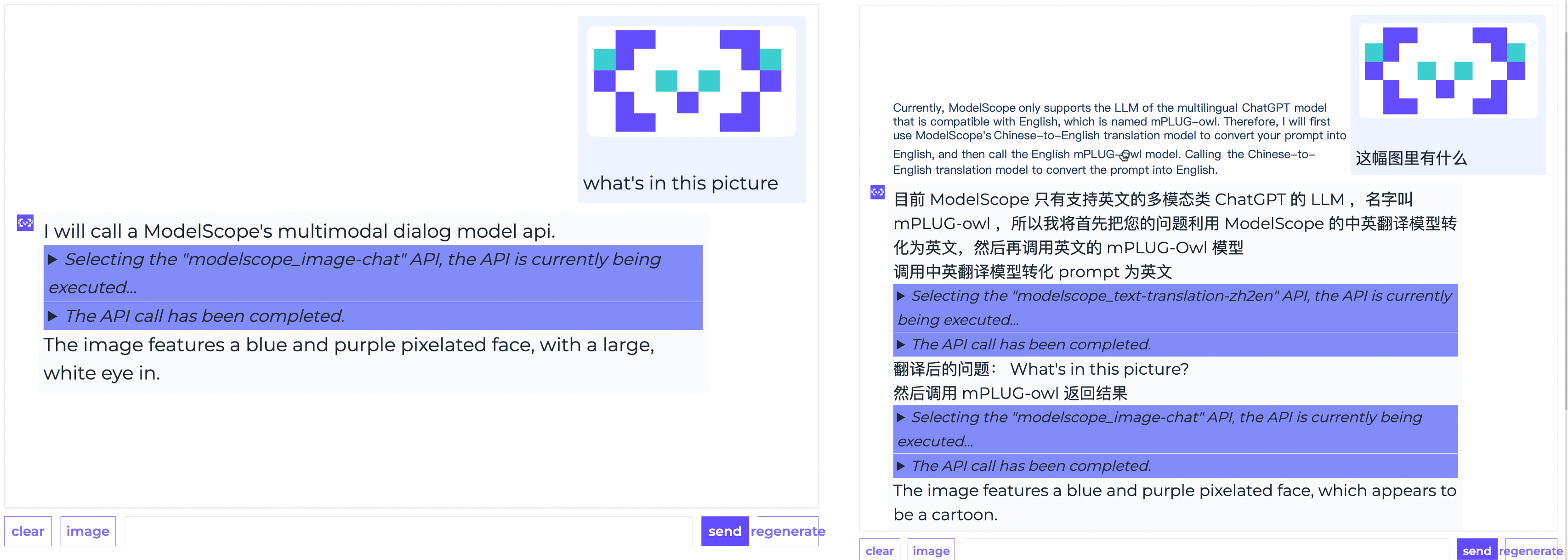}
     \caption{Single-step tool-use instructions, image-chat cases. Testing the model using the same semantic instruction in both English (left) and Chinese (right).}
     \label{fig:case_image_chat}
\end{figure*}

\begin{figure*}[!htb]
     \centering
     \includegraphics[width=\linewidth]{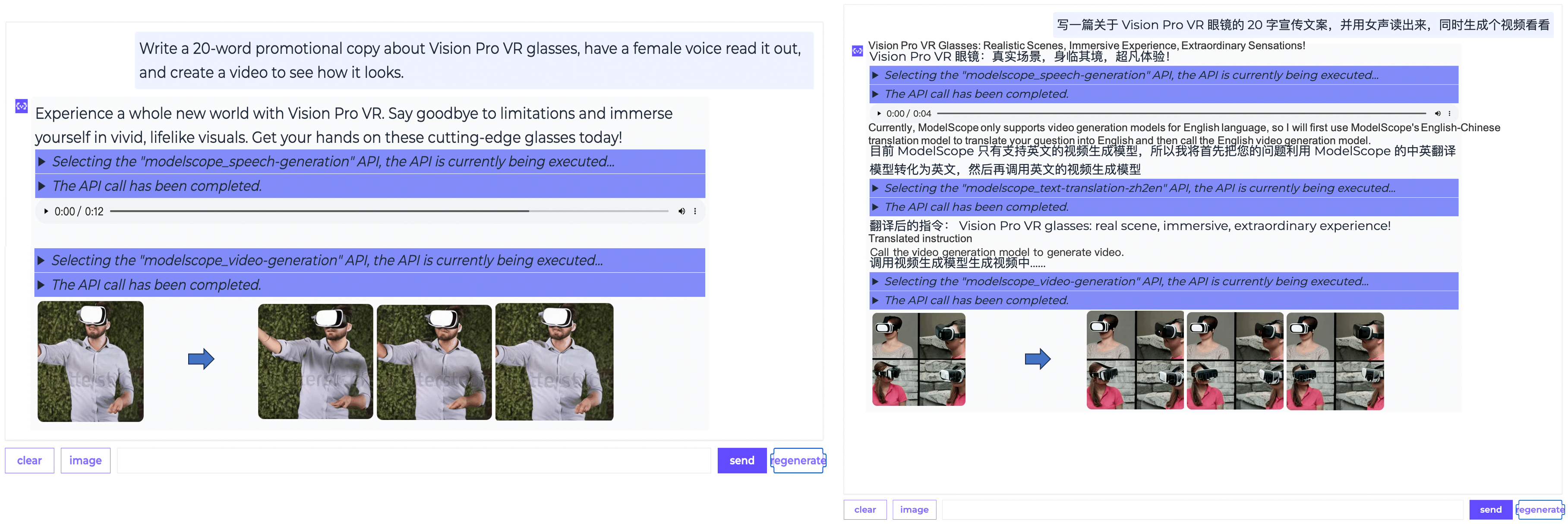}
     \caption{Multi-step tool-use instructions. We have captured a few frames of the video to display. Testing the model using the same semantic instruction in both English(left) and Chinese(right).}
     \label{fig:case_single_complex}
\end{figure*}

\begin{figure*}[!htb]
     \centering
     \includegraphics[width=\linewidth]{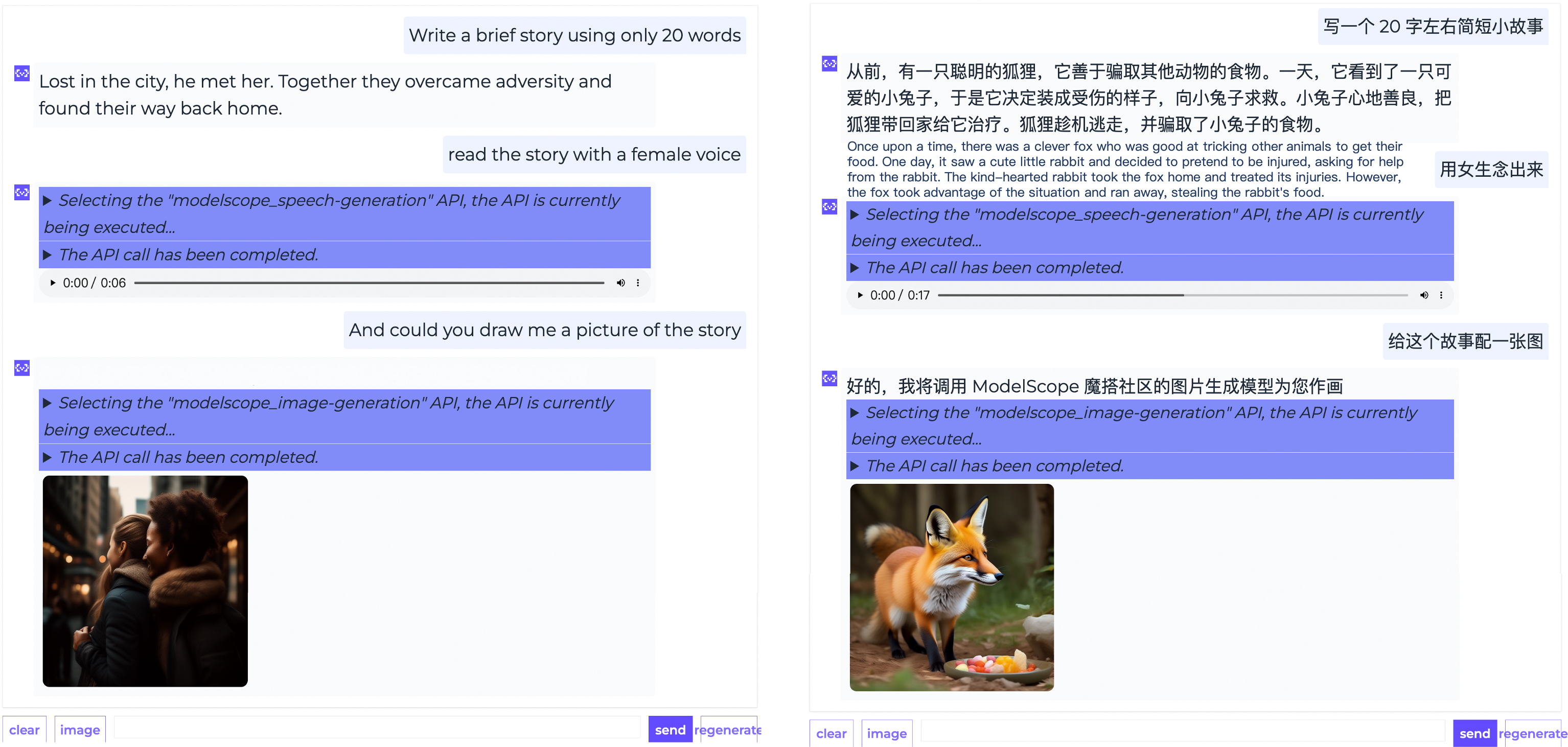}
     \caption{Multi-turn tool-use instructions, text-to-speech and text-to-image cases. Testing the model using the same semantic instruction in both English(left) and Chinese(right).}
     \label{fig:case_multi_turn_2}
\end{figure*}

\begin{figure*}[!htb]
     \centering
     \includegraphics[width=\linewidth]{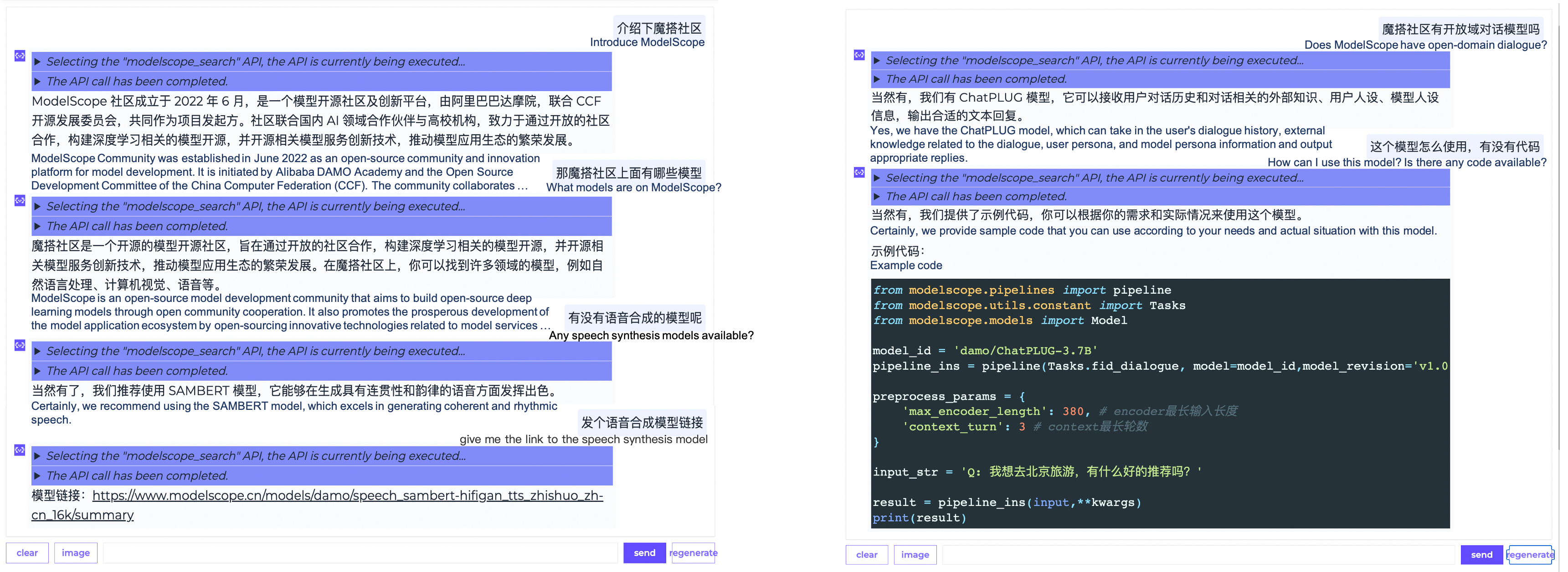}
     \caption{Multi-turn tool-use instructions, text-to-speech and text-to-image cases. Testing the model using the same semantic instruction in both English(left) and Chinese(right).}
     \label{fig:case_knowledge_qa}
\end{figure*}

\subsection{Weighted LM}
\label{sec_weighted_lm}
We give an example of the training strategy Weighted LM. As show in Figure \ref{fig: weightedlm}, tokens with different colors have different loss weights. For the user input prompt, we set the loss weight to 0, so that the model does not calculate the loss for the prompt. For the API-Agnostic text of the assistant, we keep the loss weight as 1. Finally, for the important text of the API calling, such as API name, parameters, URL, etc., we set the loss weight to 2, which can improve the generation accuracy of API calling.

\section{Cases}
\label{cases}

In this section, we show the qualitative results about ModelScopeGPT implementation based on ModelScope-Agent. 

\paragraph{Single-step Tool Use} As shown in Figure \ref{fig: case_video_generation} and \ref{fig:case_image_chat}, the instruction expects the model to generate a video and chat about the image respectively. These instructions can be completed with a single step of tool use.

\paragraph{Multi-step Tool Use} As shown in Figure \ref{fig:case_single_complex}, the instruction expects the model to write the promotional copy first, then read it, and finally generate a video. These instructions require the model to have the ability of multi-step Tool use. In the Chinese case, our model accurately completed the three-step tool use.

\paragraph{Multi-turn Tool Use} As shown in Figure \ref{fig:case_multi_turn_2}, the instruction requires the model to have the ability to multi-turn conversation and use the history conversation. Our model can accurately call the API and capture the content of the previous conversation to generate API parameters.

\paragraph{In-domain Knowledge QA} As shown in Figure \ref{fig:case_knowledge_qa}, the instruction requires the model to retrieve in-domain knowledge and use the retrieved knowledge to answer questions.

\begin{figure}[h]
     \centering
     \includegraphics[width=0.9\linewidth]{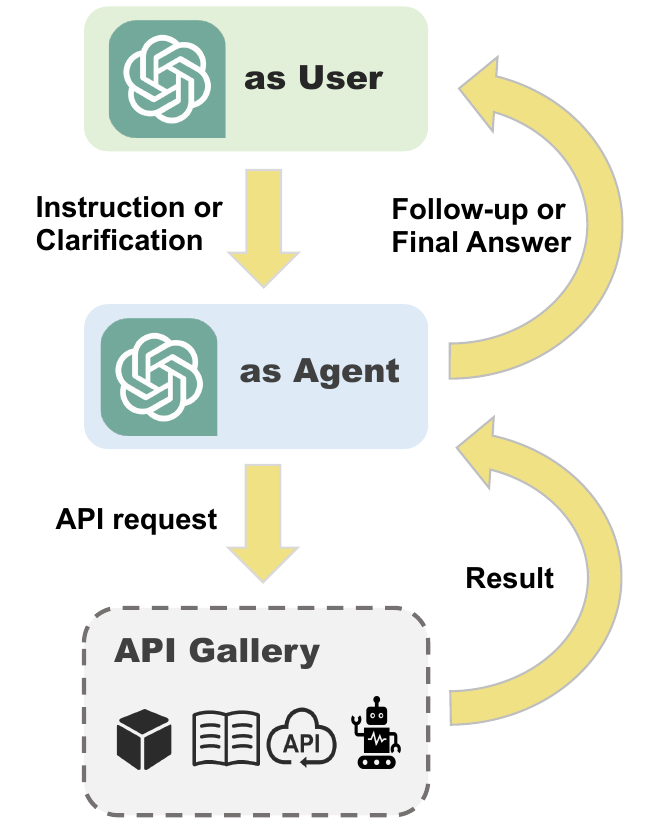}
     \caption{The data collection procedure of MSAgent-Bench.}
     \label{fig:data_generate}
     \vspace{-0.5cm}
\end{figure}

\section{Data Collection Procedure}\label{app:data_collect}

We collected our dataset by using prompt engineer to simulate the agent scenarios with two ChatGPTs (\texttt{gpt-3.5-turbo}). One of the ChatGPTs was prompted to act as the user, while the other was assigned to act as the agent. In order to expand the domains and functionalities of APIs presented in the training data, rather than the exsisting real APIs, we also included a number of synthetic APIs that were generated by ChatGPT. When these synthetic APIs were incorporated into the dialogues, we prompted another ChatGPT to serve as the API and return the relevant calling outcomes.

The data collection procedure is shown in Figure~\ref{fig:data_generate}. Initially, a set of random in-context demonstrations were provided to ChatGPT for generating an instruction. This instruction could either be a regular one or one that requires solving with APIs, depending on the demonstrations provided. Subsequently,  ChatGPT was prompt to act as an agent by first thinking about which action to undertake. If no API calls were deemed necessary, or if the user clarification is needed, the agent would respond with a follow-up response to the user. Otherwise the agent will send API request to the API gallery. After receiving the result of the API call, the agent would assess the situation and decide on the next action. This iterative process of the "user-agent-API" loop would continue until the agent determined that it was appropriate to terminate the conversation with the final answer. After acquiring the raw dataset, we applied filtering mechanisms to eliminate instances in which ChatGPT generated API requests containing hallucinated API names and parameters that were absent from the retrieved API. Additionally, we excluded instances in which ChatGPT generated illegal API requests, thus resulting in a refined and finalized dataset.

As introduced in Section~\ref{sec:dataset}, we collect instances across different languages and topics, the detailed statistics of our collected data are shown in Table~\ref{tab:dataset_statistic}.

\begin{table}[h]
 \centering
	\resizebox{0.7\linewidth}{!}{
	\begin{tabular}{lr}
		\toprule
		
		\textbf{Instance Type} & \textbf{\# Instances} \\ \hline 
             Chinese   & 532,436  \\
             English    & 66,444\\
             \hline
             Common API & 211,026\\
             Model API & 58,338\\
             API-Oriented QA & 5,000\\
             API-Agnostic Instruction& 329,776\\
            \bottomrule
	\end{tabular}
	}
        \caption{The statistics of our collected dataset.}
	\label{tab:dataset_statistic}
 \vspace{-0.3cm}
\end{table}

\section{Related Work}

\subsection{Large Language Models} 
Recent years have witnessed rapid development in the field of Large Language Models (LLMs). Typical models, such as GPT3~\citep{gpt3}, Gopher~\citep{gopher}, Chinchilla~\citep{Chinchilla}, PaLM~\citep{palm} and LLaMA~\citep{touvron2023llama}, have shown impressive zero and few-shot generalization abilities on a wide range of NLP tasks, by scaling up the model and data size. A remarkable milestone is the release of ChatGPT~\citep{chatgpt} or GPT4~\citep{gpt4}, which has greatly revolutionized the paradigm of AI development. 
As a result, a rising trend of open-source LLMs has emerged to challenge and catch up their closed-source counterparts like ChatGPT and Claude, such as BLOOM~\citep{muennighoff2022crosslingual}, LLaMA~\citep{touvron2023llama}, Falcon~\citep{almazrouei2023falcon}, ChatGLM~\citep{chatglm}. Despite the great breakthrough, LLMs are trained as text generators over plain text corpora, thus performing less well on other tasks such as multi-modal tasks. It also falls short on tasks that require up-to-date information, which are beyond the pretraining data. Using tools or external APIs can help overcome the limitations and harnesses the power of LLMs to facilitate seamless connections with downstream applications. In \modelname, we provide the whole customizable framework and best practices for building an agent system, which enables open-source LLMs to use tools and external APIs.


\subsection{Agent \& Tool Learning} 

The utilization of Large Language Models (LLMs) as a controller to construct an agent system has emerged as a prominent research area. Several related works employ prompt engineering techniques on closed-source LLMs, such as ChatGPT~\citep{chatgpt} and Claude, to enable their application in specific domains. For instance, Visual-ChatGPT~\citep{visualchatgpt} and HuggingGPT~\citep{hugginggpt} facilitate the HuggingFace model callings accessible to OpenAI LLMs. 
SayCan~\citep{saycan} and inner monologue~\citep{innermonologue} integrate LLMs with robots to achieve robotic systems. Notably, recent works such as Langchain and Auto-GPT encompass a wide range of tools, including common APIs and neural models, and enhance long-term reasoning and human-agent interaction whilst solving tasks, which demonstrate the immense potential for building a generalized agent.

Numerous endeavors have also been made to enable open-source LLMs to utilize tools. For instance, Gorilla~\citep{gorilla} and GPT4Tools~\citep{gpt4tools} generate training data using self-instruction techniques to train open-source LLMs to effectively utilize neural models. ToolAlpaca~\citep{toolalpaca} and ToolLLaMA~\citep{qin2023tool} train LLAMA using common APIs, with the distinction that ToolAlpaca employs synthetic APIs from LLMS, whereas ToolLLaMA utilizes real APIs.

Overall, compared to the above-mentioned methods, \modelname differs in the following aspects. Firstly, our method includes a universal training framework that supports user-customized agent learning for open-source models to meet industrial needs. Secondly, \modelname can support various APIs in different fields, including model APIs and common APIs, while previous works only support certain specific APIs.

\end{document}